\documentclass[10pt,twocolumn,letterpaper]{article}

\usepackage{cvpr}
\usepackage{times}
\usepackage{epsfig}
\usepackage{graphicx}
\usepackage{amsmath}
\usepackage{amssymb}
\usepackage{bbm}
\usepackage{multirow}
\usepackage[linesnumbered,ruled,vlined]{algorithm2e}
\DontPrintSemicolon 
\SetKwInput{KwInput}{Input}                
\SetKwInput{KwOutput}{Output}
\usepackage[dvipsnames]{xcolor}

\usepackage[pagebackref=true,breaklinks=true,letterpaper=true,colorlinks,bookmarks=false]{hyperref}

\cvprfinalcopy 

\def\cvprPaperID{1320} 

\ifcvprfinal\fi
\begin{document}

\title{Point-GNN: Graph Neural Network for 3D Object Detection in a Point Cloud}

\author{Weijing Shi and Ragunathan (Raj) Rajkumar\\
Carnegie Mellon University\\
Pittsburgh, PA 15213\\
{\tt\small \{weijings, rajkumar\}@cmu.edu}
}

\maketitle

\begin{abstract}
In this paper, we propose a graph neural network to detect objects from a LiDAR point cloud. Towards this end, we encode the point cloud efficiently in a fixed radius near-neighbors graph. We design a graph neural network, named Point-GNN, to predict the category and shape of the object that each vertex in the graph belongs to. In Point-GNN, we propose an auto-registration mechanism to reduce translation variance, and also design a box merging and scoring operation to combine detections from multiple vertices accurately. Our experiments on the KITTI benchmark show the proposed approach achieves leading accuracy using the point cloud alone and can even surpass fusion-based algorithms. Our results demonstrate the potential of using the graph neural network as a new approach for 3D object detection.
The code is available at \href{https://github.com/WeijingShi/Point-GNN}{https://github.com/WeijingShi/Point-GNN}. 
\end{abstract}

\section{Introduction}

Understanding the 3D environment is vital in robotic perception. A point cloud that composes a set of points in space is a widely-used format for 3D sensors such as LiDAR. Detecting objects accurately from a point cloud is crucial in applications such as autonomous driving.

Convolutional neural networks that detect objects from images rely on the convolution operation. While the convolution operation is efficient, it requires a regular grid as input. Unlike an image, a point cloud is typically sparse and not spaced evenly on a regular grid. Placing a point cloud on a regular grid generates an uneven number of points in the grid cells. Applying the same convolution operation on such a grid leads to potential information loss in the crowded cells or wasted computation in the empty cells.

Recent breakthroughs in using neural networks \cite{PointNet} \cite{deepset} allow an unordered set of points as input. Studies take advantage of this type of neural network to extract point cloud features without mapping the point cloud to a grid. However, they typically need to sample and group points iteratively to create a \textit{point set} representation. The repeated grouping and sampling on a large point cloud can be computationally costly. Recent 3D detection approaches \cite{PointPillars}\cite{STD}\cite{PointRCNN} often take a hybrid approach to use a grid and a set representation in different stages. Although they show some promising results,  such hybrid strategies may suffer the shortcomings of both representations.

In this work, we propose to use a graph as a compact representation of a point cloud and design a graph neural network called Point-GNN to detect objects. We encode the point cloud natively in a graph by using the points as the graph vertices. The edges of the graph connect neighborhood points that lie within a fixed radius, which allows feature information to flow between neighbors. Such a graph representation adapts to the structure of a point cloud directly without the need to make it regular. A graph neural network reuses the graph edges in every layer, and avoids grouping and sampling the points repeatedly. 

\begin{figure}[t]
\centering
\includegraphics[width=\linewidth]{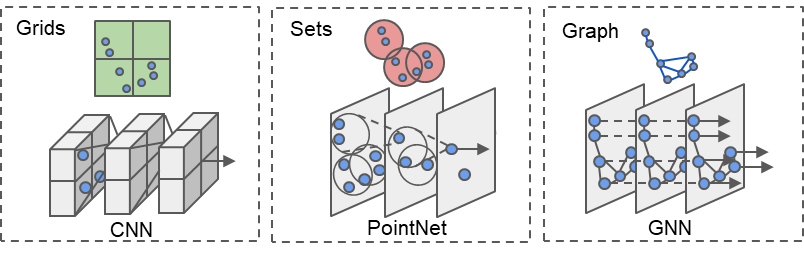}
  \caption{Three point cloud representations and their common processing methods.}
\label{fig:representation}
\vspace{-2mm}
\end{figure}

Studies \cite{3DGNN} \cite{Landrieu_2018_CVPR} \cite{gnn_classification} \cite{dgcnn} have looked into using graph neural network for the classification and the semantic segmentation of a point cloud. However, little research has looked into using a graph neural network for the 3D object detection in a point cloud. Our work demonstrates the feasibility of using a GNN for highly accurate object detection in a point cloud.

Our proposed graph neural network Point-GNN takes the point graph as its input. It outputs the category and bounding boxes of the objects to which each vertex belongs. Point-GNN is a one-stage detection method that detects multiple objects in a single shot. To reduce the translation variance in a graph neural network, we introduce an auto-registration mechanism which allows points to align their coordinates based on their features. We further design a box merging and scoring operation to combine detection results from multiple vertices accurately.

We evaluate the proposed method on the KITTI benchmark. On the KITTI benchmark, Point-GNN achieves the state-of-the-art accuracy using the point cloud alone and even surpasses sensor fusion approaches. Our Point-GNN shows the potential of a new type 3D object detection approach using graph neural network, and it can serve as a strong baseline for the future research. We conduct an extensive ablation study on the effectiveness of the components in Point-GNN. 

In summery, the contributions of this paper are:
\begin{itemize}
  \item We propose a new object detection approach using graph neural network on the point cloud.  
  \item We design Point-GNN, a graph neural network with an auto-registration mechanism that detects multiple objects in a single shot. 
  \item We achieve state-of-the-art 3D object detection accuracy in the KITTI benchmark and analyze the effectiveness of each component in depth.  
\end{itemize}

\section{Related Work}
Prior work in this context can be grouped into three categories, as shown in Figure \ref{fig:representation}.

\noindent\textbf{Point cloud in grids.} Many recent studies convert a point cloud to a regular grid to utilize convolutional neural networks. \cite{PIXOR} projects a point cloud to a 2D Bird's Eye View (BEV) image and uses a 2D CNN for object detection. \cite{MV3D} projects a point cloud to both a BEV image and a Front View (FV) image before applying a 2D CNN on both. Such projection induces a quantization error due to the limited image resolution. Some approaches keep a point cloud in 3D coordinates.  \cite{VoxelNet} represents points in 3D voxels and applies 3D convolution for object detection. When the resolution of the voxels grows, the computation cost of 3D CNN grows cubically, but many voxels are empty due to point sparsity. Optimizations such as the sparse convolution \cite{SECOND} reduce the computation cost. Converting a point cloud to a 2D/3D grid suffers from the mismatch between the irregular distribution of points and the regular structure of the grids. 

\noindent\textbf{Point cloud in sets.}
Deep learning techniques on sets such as PointNet \cite{PointNet} and DeepSet\cite{deepset} show neural networks can extract features from an unordered set of points directly. In such a method, each point is processed by a multi-layer perceptron (MLP) to obtain a point feature vector. Those features are aggregated by an average or max pooling function to form a global feature vector of the whole set. \cite{PointNetPlusPlus} further proposes the hierarchical aggregation of point features, and generates local subsets of points by sampling around some key points. The features of those subsets are then again grouped into sets for further feature extraction. Many 3D object detection approaches take advantage of such neural networks to process a point cloud without mapping it to a grid. However, the sampling and grouping of points on a large scale lead to additional computational costs. Most object detection studies only use the neural network on sets as a part of the pipeline. \cite{F-PointNet} generates object proposals from camera images and uses \cite{PointNetPlusPlus} to separate points that belong to an object from the background and predict a bounding box. \cite{PointRCNN} uses  \cite{PointNetPlusPlus} as a backbone network to generate bounding box proposals directly from a point cloud. Then, it uses a second-stage point network to refine the bounding boxes. Hybrid approaches such as \cite{VoxelNet} \cite{SECOND} \cite{PointPillars} \cite{STD} use \cite{PointNet} to extract features from local point sets and place the features on a regular grid for the convolutional operation. Although they reduce the local irregularity of the point cloud to some degree, they still suffer the mismatch between a regular grid and the overall point cloud structure. 

\noindent\textbf{Point cloud in graphs.}
Research on graph neural network \cite{gnn_survey}  seeks to generalize the convolutional neural network to a graph representation. A GNN iteratively updates its vertex features by aggregating features along the edges. Although the aggregation scheme sometimes is similar to that in deep learning on sets, a GNN allows more complex features to be determined along the edges. It typically does not need to sample and group vertices repeatedly. In the computer vision domain, a few approaches represent the point cloud as a graph. \cite{3DGNN} uses a recurrent GNN for the semantic segmentation on RGBD data. \cite{Landrieu_2018_CVPR} partitions a point cloud to simple geometrical shapes and link them into a graph for semantic segmentation. \cite{gnn_classification} \cite{dgcnn} look into classifying a point cloud using a GNN. So far, few investigations have looked into designing a graph neural network for object detection, where an explicit prediction of the object shape is required. 

Our work differs from previous work by designing a GNN for object detection. Instead of converting a point cloud to a regular gird, such as an image or a voxel, we use a graph representation to preserve the irregularity of a point cloud. Unlike the techniques that sample and group the points into sets repeatedly, we construct the graph \textit{once}. The proposed Point-GNN then extracts features of the point cloud by iteratively updating vertex features on the same graph.  Our work is a single-stage detection method without the need to develop a second-stage refinement neural networks like those in \cite{MV3D}\cite{PointRCNN}\cite{STD}\cite{UberATG-MMF}\cite{F-PointNet}. 

\begin{figure*}
\centering
\includegraphics[width=0.98\linewidth]{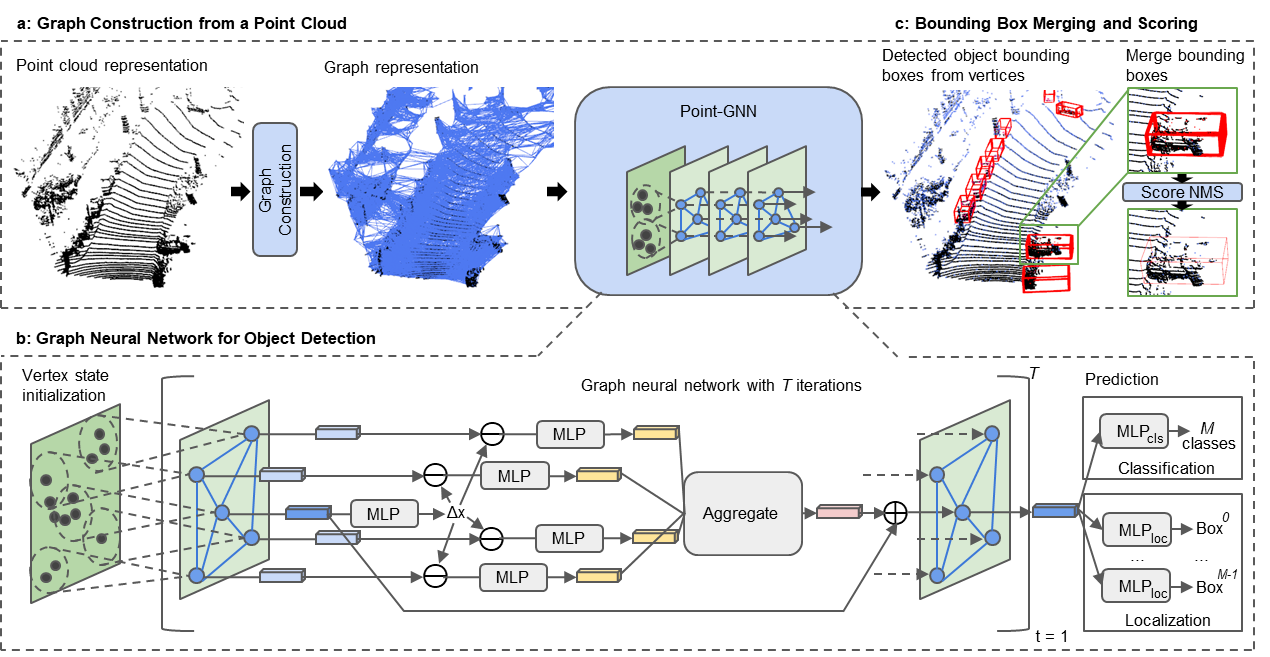}
   \caption{The architecture of the proposed approach. It has three main components: (a) graph construction from a point cloud, (b) a graph neural network for object detection, and (c) bounding box merging and scoring.}
\label{fig:architecture}

\end{figure*}

\section{Point-GNN for 3D Object Detection in a Point Cloud}

In this section, we describe the proposed approach to detect 3D objects from a point cloud. As shown in Figure \ref{fig:architecture}, the overall architecture of our method contains three components: (a) graph construction, (b) a GNN of $T$ iterations, and (c) bounding box merging and scoring. 
 
\subsection{Graph Construction}
Formally, we define a point cloud of $N$ points as a set $P = \{p_1, ..., p_N\}$, where $p_i = (x_i, s_i)$ is a point with both 3D coordinates  $x_i \in \mathbb{R}^3 $ and the state value $s_i \in \mathbb{R}^k $ a $k$-length vector that represents the point property. The state value $s_i$ can be the reflected laser intensity or the features which encode the surrounding objects. Given a point cloud $P$, we construct a graph $G = (P, E)$ by using $P$ as the vertices and connecting a point to its neighbors within a fixed radius $r$, i.e.
\begin{equation}\label{eq:graph}
    E = \{(p_i, p_j)\ |\ {\Vert x_i - x_j \Vert}_2 < r\}   
\end{equation}

The construction of such a graph is the well-known fixed radius near-neighbors search problem. By using a cell list to find point pairs that are within a given cut-off distance, we can efficiently solve the problem with a runtime complexity of $O(cN)$ where $c$ is the max number of neighbors within the radius \cite{BENTLEY1977209}.

In practice, a point cloud commonly comprises tens of thousands of points. Constructing a graph with all the points as vertices imposes a substantial computational burden. Therefore, we use a voxel downsampled point cloud $\hat{P}$ for the graph construction.  It must be noted that the voxels here are only used to reduce the density of a point cloud and they are not used as the representation of the point cloud. We still use a graph to present the downsampled point cloud. To preserve the information within the original point cloud, we encode the dense point cloud in the initial state value $s_i$ of the vertex. More specifically,  we search the raw points within a $r_0$ radius of each vertex and use the neural network on sets to extract their features. We follow ~\cite{PointPillars} ~\cite{VoxelNet} and embed the lidar reflection intensity and the relative coordinates using an $MLP$ and then aggregate them by the $Max$ function. We use the resulting features as the initial state value of the vertex. After the graph construction, we process the graph with a GNN, as shown in Figure \ref{fig:architecture}b.

\subsection{Graph Neural Network with Auto-Registration}
A typical graph neural network refines the vertex features by aggregating features along the edges. In the $(t+1)^{th}$ iteration, it updates each vertex feature in the form:
\begin{equation}\label{eq:gnn_general}
    \begin{split}
        v_i^{t+1} &= g^t(\rho(\{e_{ij}^{t}\ |\ (i, j) \in E \}), v_i^{t}) \\ 
        e_{ij}^{t} &= f^t(v_i^{t}, v_j^{t}) \\ 
    \end{split}
\end{equation}
where $e^t$ and $v^t$ are the edge and vertex features from the $t^{th}$ iteration. A function $f^t(.)$ computes the edge feature between two vertices. $\rho(.)$ is a set function which aggregates the edge features for each vertex. $g^t(.)$ takes the aggregated edge features to update the vertex features. The graph neural network then outputs the vertex features or repeats the process in the next iteration. 

In the case of object detection, we design the GNN to refine a vertex's state to include information about the object where the vertex belongs. Towards this goal, we re-write Equation (\ref{eq:gnn_general}) to refine a vertex's state using its neighbors' states:
\begin{equation}\label{eq:gnn_pcl}
    s_i^{t+1} = g^t(\rho(\{f^t(x_j - x_i, s_j^{t})\ |\ (i, j) \in E\}), s_i^t)
\end{equation} 
Note that we use the relative coordinates of the neighbors as input to $f^t(.)$ for the edge feature extraction. The relative coordinates induce translation invariance against the global shift of the point cloud.  However, it is still sensitive to translation within the neighborhood area. When a small translation is added to a vertex, the local structure of its neighbors remains similar. But the relative coordinates of the neighbors are all changed, which increases the input variance to $f^t(.)$. To reduce the translation variance, we propose aligning neighbors' coordinates by their structural features instead of the center vertex coordinates. Because the center vertex already contains some structural features from the previous iteration,  we can use it to predict an alignment offset, and propose an \textit{auto-registration} mechanism:
\begin{equation}\label{eq:gnn_invariance}
    \begin{split}
        {\Delta x_i}^t &= h^t(s_i^{t})\\
        s_i^{t+1} &= g^t(\rho(\{f(x_j-x_i+{\Delta x_i}^t, s_j^{t})\}, s_i^t)
    \end{split}
\end{equation}    
${\Delta x_i^t}$ is the coordination offset for the vertices to register their coordinates. $h^t(.)$ calculates the offset using the center vertex state value from the previous iteration. By setting $h^t(.)$ to output zero, the GNN can disable the offset if necessary. In that case, the GNN returns to Equation (\ref{eq:gnn_pcl}). We analyze the effectiveness of this auto-registration mechanism in Section \ref{Experiments}. 

As shown in Figure \ref{fig:architecture}b, we model $f^t(.)$, $g^t(.)$ and $h^t(.)$ using multi-layer perceptrons ($MLP$) and add a residual connection in $g^t(.)$. We choose $\rho(.)$ to be $Max$ for its robustness\cite{PointNet}. A single iteration in the proposed graph network is then given by:
\begin{equation}\label{eq:gnn_point}
    \begin{split}
	 {\Delta x_i}^t &= MLP_h^t(s_i^{t})\\
	e_{ij}^{t} &= MLP_f^t([x_j-x_i+{\Delta x_i}^t, s_j^{t}])\\
        s_i^{t+1} &= MLP_g^t(Max(\{e_{ij}\ | (i,j)\in E\})) + s_i^t 
    \end{split}
\end{equation}
where $[,]$ represents the concatenation operation. 

Every iteration $t$ uses a different set of $MLP^t$, which is not shared among iterations. 
After $T$ iterations of the graph neural network, we use the vertex state value to predict both the category and the bounding box of the object where the vertex belongs. A classification branch $MLP_{cls}$ computes a multi-class probability. Finally, a localization branch $MLP_{loc}$ computes a bounding box for each 
 class. 
 
\subsection{Loss}
For the object category, the classification branch computes a multi-class probability distribution $\{p_{c_1}, ..., p_{c_M}\}$ for each vertex. $M$ is the total number of object classes, including the \textit{Background} class. If a vertex is within a bounding box of an object,  we assign the object class to the vertex. If a vertex is outside any bounding boxes, we assign the background class to it.  We use the average cross-entropy loss as the classification loss.
\begin{equation}\label{eq:classification}
    l_{cls} = -\frac{1}{N}\sum_{i=1}^N\sum_{j=1}^{M} y_{c_j}^i log (p_{c_j}^i)
\end{equation} 
where $p_c^i$ and $y_c^i$ are the predicted probability and the one-hot class label for the $i$-th vertex respectively. 

For the object bounding box, we predict it in the 7 degree-of-freedom format $b=(x, y, z, l, h, w, \theta)$, where $(x, y, z)$ represent the center position of the bounding box, $(l, h, w)$ represent the box length, height and width respectively, and $\theta$ is the yaw angle. We encode the bounding box with the vertex coordinates $(x_v, y_v, z_v)$ as follows:
\begin{equation}\label{eq:classification}
    \begin{split}
        \delta_x &= \frac{x - x_v}{l_m},\ \delta_y = \frac{y - y_v}{h_m},\ \delta_z = \frac{z - z_v}{w_m} \\
        \delta_l &= log(\frac{l}{l_m}),\ \delta_h = log(\frac{h}{h_m}),\ \delta_w = log(\frac{w}{w_m}) \\
       \delta_\theta &= \frac{\theta-\theta_0}{\theta_m}
    \end{split}
\end{equation}  
where $l_m, h_m, w_m, \theta_0, \theta_m$ are constant scale factors.

The localization branch predicts the encoded bounding box $\delta_b = ( \delta_x,  \delta_y,  \delta_z,  \delta_l,  \delta_h,  \delta_w,  \delta_\theta)$ for each class. If a vertex is within a bounding box, we compute the Huber loss \cite{huber1964} between the ground truth and our prediction. If a vertex is outside any bounding boxes or it belongs to a class that we do not need to localize, we set its localization loss as zero. We then average the localization loss of all the vertices: 
\begin{equation}\label{eq:localization}
    l_{loc} = \frac{1}{N}\sum_{i=1}^N\mathbbm{1}(v_i \in b_{interest}) \sum_{\delta \in \delta_{b_i}} l_{huber}(\delta-\delta^{gt})
\end{equation} 

To prevent over-fitting, we add $L1$ regularization to each MLP. The total loss is then:
 \begin{equation}\label{eq:total_loss}
    l_{total} = \alpha l_{cls} + \beta l_{loc} + \gamma l_{reg}
\end{equation} 
where $\alpha$, $\beta$ and $\gamma$ are constant weights to balance each loss.  

\subsection{Box Merging and Scoring}
As multiple vertices can be on the same object, the neural network can output multiple bounding boxes of the same object. It is necessary to merge these bounding boxes into  one and also assign a confidence score. Non-maximum suppression (NMS) has been widely used for this purpose. The common practice is to select the box with the highest classification score and suppress the other overlapping boxes. However, the classification score does not always reflect the localization quality. Notably, a partially occluded object can have a strong clue indicating the type of the object but lacks enough shape information. The standard NMS can pick an inaccurate bounding box base on the classification score alone.

\begin{algorithm} \label{nms}
\SetAlgoLined
\KwInput{$\mathcal{B} = \{b_1, ..., b_n\}$, $\mathcal{D} = \{d_1, ..., d_n\}$, $T_h$\\
$\mathcal{B}$ is the set of detected bounding boxes.\\
$\mathcal{D}$ is the corresponding detection scores.\\
$T_h$ is an overlapping threshold value.\\
\textcolor{ForestGreen}{Green} color marks the main modifications. 
}
 $\mathcal{M}\ {\leftarrow}\ \{ \}$,
 $\mathcal{Z}\ {\leftarrow}\ \{ \}$ \;
 \While{$\mathcal{B}\neq empty$}{
  $ i {\leftarrow} argmax\ {D}$\;
  $\mathcal{L}\ {\leftarrow}\ \{ \}$\;
  \For{$b_j$ in $\mathcal{B}$ }{
    \If{$iou(b_i, b_j) > T_h$} {
        $\mathcal{L} {\leftarrow}\mathcal{L} \cup b_j$\;
        $\mathcal{B} {\leftarrow}\mathcal{B} - b_j$,
        $\mathcal{D} {\leftarrow}\mathcal{D} - d_j$\;
    }
  }
  \textcolor{ForestGreen}{$m\ {\leftarrow}\ median(\mathcal{L})$}\;
  \textcolor{ForestGreen}{$o\ {\leftarrow}\ occlusion(m)$}\;
  \textcolor{ForestGreen}{$z\ {\leftarrow}\ (o+1)\sum_{b_k\in \mathcal{L}} IoU(m, b_k)d_k$}\;
  $\mathcal{M} {\leftarrow}\mathcal{M} \cup m$, 
  $\mathcal{Z} {\leftarrow}\mathcal{Z} \cup z$\;
 }
 \Return{$\mathcal{M}$, $\mathcal{Z}$}
 \caption{NMS with Box Merging and Scoring}
\end{algorithm}

To improve the localization accuracy, we propose to calculate the merged box by considering the entire overlapped box cluster. More specifically, we consider the median position and size of the overlapped bounding boxes.  We also compute the confidence score as the sum of the classification scores weighted by the Intersection-of-Union (IoU) factor and an occlusion factor. The occlusion factor represents the occupied volume ratio. Given a box $b_i$, let $l_i$, $w_i$, $h_i$ be its length, width and height, and let $v_i^l$, $v_i^w$, $v_i^h$ be the unit vectors that indicate their directions respectively. $x_j$ are the coordinates of point $p_j$. The occlusion factor $o_i$ is then:
\begin{equation}\label{eq:coverage_box}
    o_i = \frac{1}{l_i w_i h_i}\prod_{v \in \{v_i^l, v_i^w, v_i^h\}}\max_{p_j \in b_i} (v^Tx_j) - \min_{p_j \in b_i}(v^Tx_j) \\
\end{equation}

We modify standard NMS as shown in Algorithm \ref{nms}. It returns the merged bounding boxes $\mathcal{M}$ and their confidence score $\mathcal{Z}$. We will study its effectiveness in Section \ref{Experiments}.
 
\begin{table*}[]
\centering
\resizebox{\linewidth}{!}{
\begin{tabular}{c|c|ccc|ccc|ccc}
\hline
\multirow{2}{*}{Method} & \multirow{2}{*}{Modality} & \multicolumn{3}{c|}{Car}                         & \multicolumn{3}{c|}{Pedestrian}                  & \multicolumn{3}{c}{Cyclist}                      \\
                        &                           & Easy           & Moderate       & Hard           & Easy           & Moderate       & Hard           & Easy           & Moderate       & Hard           \\ \hline\hline
UberATG-ContFuse\cite{UberATG-ContFuse}        & LiDAR + Image             & 82.54          & 66.22          & 64.04          & N/A            & N/A            & N/A            & N/A            & N/A            & N/A            \\
AVOD-FPN\cite{AVOD}                & LiDAR + Image             & 81.94          & 71.88          & 66.38          & 50.80           & 42.81          & 40.88          & 64.00             & 52.18          & 46.61          \\
F-PointNet\cite{F-PointNet}              & LiDAR + Image             & 81.20           & 70.39          & 62.19          & 51.21          & 44.89          & 40.23          & 71.96          & 56.77          & 50.39          \\
UberATG-MMF\cite{UberATG-MMF}             & LiDAR + Image             & 86.81          & 76.75          & 68.41          & N/A            & N/A            & N/A            & N/A            & N/A            & N/A            \\ \hline
VoxelNet\cite{VoxelNet}                & LiDAR                     & 81.97          & 65.46          & 62.85          & \textbf{57.86} & \textbf{53.42} & \textbf{48.87} & 67.17          & 47.65          & 45.11          \\
SECOND\cite{SECOND}                  & LiDAR                     & 83.13          & 73.66          & 66.20           & 51.07          & 42.56          & 37.29          & 70.51          & 53.85          & 53.85          \\
PointPillars\cite{PointPillars}            & LiDAR                     & 79.05          & 74.99          & 68.30           & 52.08          & 43.53          & 41.49          & 75.78          & 59.07          & 52.92          \\
PointRCNN\cite{PointRCNN}               & LiDAR                     & 85.94          & 75.76          & 68.32          & 49.43          & 41.78          & 38.63          & 73.93          & 59.60           & 53.59          \\
STD\cite{STD}                     & LiDAR                     & 86.61          & 77.63          & \textbf{76.06} & 53.08          & 44.24          & 41.97          & \textbf{78.89} & 62.53          & 55.77          \\ \hline
\textbf{Our Point-GNN}                    & LiDAR                     & \textbf{88.33} & \textbf{79.47} & 72.29          & 51.92          & 43.77          & 40.14          & 78.60           & \textbf{63.48} & \textbf{57.08} \\ \hline
\end{tabular}
}
\vspace*{0mm}
\caption{The Average Precision (AP) comparison of 3D object detection on the KITTI \textit{test} dataset.}
\label{3D_table}
\end{table*}

\begin{table*}[]
\centering
\resizebox{\linewidth}{!}{
\begin{tabular}{ccccccccccc}
\hline
\multicolumn{1}{c|}{\multirow{2}{*}{Method}} & \multicolumn{1}{c|}{\multirow{2}{*}{Modality}} & \multicolumn{3}{c|}{Car}                                                          & \multicolumn{3}{c|}{Pedestrian}                                                   & \multicolumn{3}{c}{Cyclist}                                        \\
\multicolumn{1}{c|}{}                        & \multicolumn{1}{c|}{}                          & Easy                 & Moderate             & \multicolumn{1}{c|}{Hard}           & Easy                 & Moderate             & \multicolumn{1}{c|}{Hard}           & Easy                 & Moderate             & Hard                 \\ \hline\hline
\multicolumn{1}{c|}{UberATG-ContFuse\cite{UberATG-ContFuse}}        & \multicolumn{1}{c|}{LiDAR + Image}             & 88.81                & 85.83                & \multicolumn{1}{c|}{77.33}          & N/A                  & N/A                  & \multicolumn{1}{c|}{N/A}            & N/A                  & N/A                  & N/A                  \\
\multicolumn{1}{c|}{AVOD-FPN\cite{AVOD}}                & \multicolumn{1}{c|}{LiDAR + Image}             & 88.53                & 83.79                & \multicolumn{1}{c|}{77.9}           & 58.75                & 51.05                & \multicolumn{1}{c|}{47.54}          & 68.06                & 57.48                & 50.77                \\
\multicolumn{1}{c|}{F-PointNet\cite{F-PointNet}}              & \multicolumn{1}{c|}{LiDAR + Image}             & 88.70                 & 84 .00                  & \multicolumn{1}{c|}{75.33}          & 58.09                & 50.22                & \multicolumn{1}{c|}{47.20}           & 75.38                & 61.96                & 54.68                \\
\multicolumn{1}{c|}{UberATG-MMF\cite{UberATG-MMF} }            & \multicolumn{1}{c|}{LiDAR + Image}             & 89.49                & 87.47                & \multicolumn{1}{c|}{79.10}           & N/A                  & N/A                  & \multicolumn{1}{c|}{N/A}            & N/A                  & N/A                  & N/A                  \\ \hline
\multicolumn{1}{c|}{VoxelNet\cite{VoxelNet}}                & \multicolumn{1}{c|}{LiDAR}                     & 89.60                 & 84.81                & \multicolumn{1}{c|}{78.57}          & \textbf{65.95}       & \textbf{61.05}       & \multicolumn{1}{c|}{\textbf{56.98}} & 74.41                & 52.18                & 50.49                \\
\multicolumn{1}{c|}{SECOND\cite{SECOND}}                  & \multicolumn{1}{c|}{LiDAR}                     & 88.07                & 79.37                & \multicolumn{1}{c|}{77.95}          & 55.10                 & 46.27                & \multicolumn{1}{c|}{44.76}          & 73.67                & 56.04                & 48.78                \\
\multicolumn{1}{c|}{PointPillars\cite{PointPillars}}            & \multicolumn{1}{c|}{LiDAR}                     & 88.35                & 86.10                 & \multicolumn{1}{c|}{79.83}          & 58.66                & 50.23                & \multicolumn{1}{c|}{47.19}          & 79.14                & 62.25                & 56.00                   \\
\multicolumn{1}{c|}{STD\cite{STD}}                     & \multicolumn{1}{c|}{LiDAR}                     & 89.66                & 87.76                & \multicolumn{1}{c|}{\textbf{86.89}} & 60.99                & 51.39                & \multicolumn{1}{c|}{45.89}          & 81.04                & 65.32                & 57.85                \\ \hline
\multicolumn{1}{c|}{\textbf{Our Point-GNN}}                    & \multicolumn{1}{c|}{LiDAR}                     & \textbf{93.11}       & \textbf{89.17}       & \multicolumn{1}{c|}{83.9}           & 55.36                & 47.07                & \multicolumn{1}{c|}{44.61}          & \textbf{81.17}       & \textbf{67.28}       & \textbf{59.67}       \\ \hline
\end{tabular}
}
\vspace*{0mm}
\caption{The Average Precision (AP) comparison of Bird's Eye View (BEV) object detection on the KITTI \textit{test} dataset.}
\label{BEV_table}
\end{table*}

\begin{figure*}
\begin{center}
\includegraphics[width=0.90\linewidth]{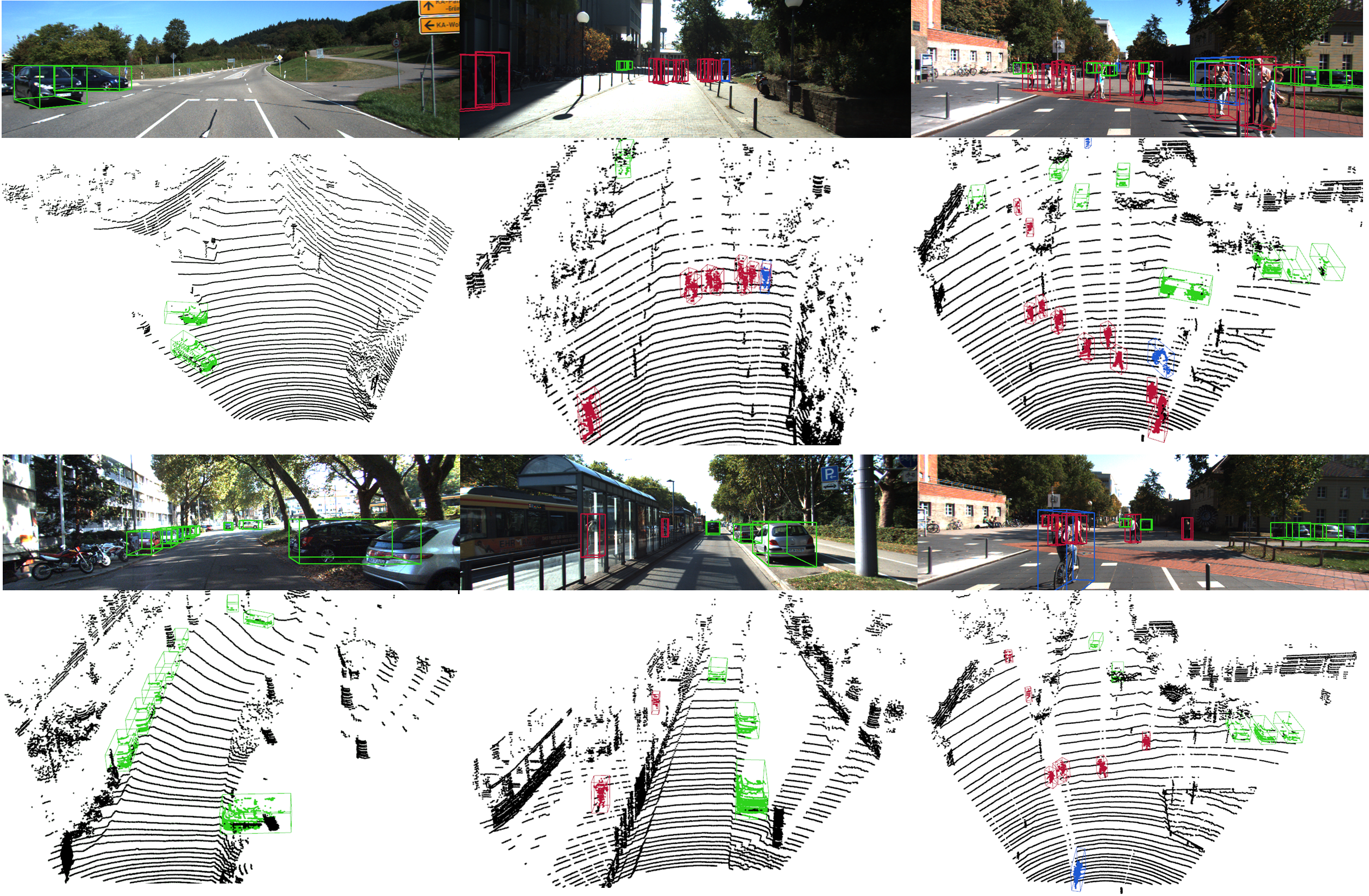}
\end{center}
   \caption{Qualitative results on the KITTI \textit{test} dataset using Point-GNN. We show the predicted 3D bounding box of Cars (green), Pedestrians (red) and Cyclists (blue) on both the image and the point cloud. Best viewed in color.}
\label{fig:quality}
\end{figure*}

\section{Experiments}\label{Experiments}
\subsection{Dataset}
We evaluate our design using the widely used KITTI object detection benchmark \cite{Geiger2012CVPR}. The KITTI dataset contains 7481 training samples and 7518 testing samples. Each sample provides both the point cloud and the camera image. We only use the point cloud in our approach. Since the dataset only annotates objects that are visible within the image, we process the point cloud only within the field of view of the image. The KITTI benchmark evaluates the average precision (AP) of three types of objects: Car, Pedestrian and Cyclist. Due to the scale difference, we follow the common practice \cite{PointPillars}\cite{VoxelNet}\cite{SECOND}\cite{STD} and train one network for the Car and another network for the Pedestrian and Cyclist.  For training, we remove samples that do not contain objects of interest.  

\subsection{Implementation Details}
We use three iterations ($T=3$) in our graph neural network. During training, we limit the maximum number of input edges per vertex to 256. During inference, we use all the input edges. All GNN layers perform auto-registration using a two-layer $MLP_h$ of units $(64,3)$. The $MLP_{cls}$ is of size $(64, \#(classes))$. For each class, $MLP_{loc}$ is of size $(64, 64, 7)$. 

\noindent \textbf{Car:}  We set $(l_m, h_m, w_m)$ to the median size of Car bounding boxes $(3.88m, 1.5m, 1.63m)$. We treat a side-view car with $\theta \in [-\pi/4, \pi/4]$ and a front-view car $\theta \in [\pi/4, 3\pi/4]$ as two different classes. Therefore, we set $\theta_0=0$ and $\theta_0=\pi/2$ respectively. The scale $\theta_m$ is set as $\pi/2$. Together with the \textit{Background} class and \textit{DoNotCare} class, 4 classes are predicted.  We construct the graph with $r=4m$ and $r_0=1m$. We set $\hat{P}$ as a downsampled point cloud by a voxel size of 0.8 meters in training and 0.4 meters in inference.  $MLP_f$ and $MLP_g$, are both of sizes $(300, 300)$. For the initial vertex state, we use an $MLP$ of $(32, 64, 128, 300)$ for embedding raw points and another $MLP$ of $(300, 300)$ after the $Max$ aggregation. We set $T_h = 0.01$ in NMS. 

\noindent \textbf{Pedestrian and Cyclist.}  Again, we set $(l_m, h_m, w_m)$ to the median bounding box size. We set $(0.88m, 1.77m, 0.65m)$ for Pedestrian and $(1.76m, 1.75m, 0.6m)$ for Cyclist. Similar to what we did with the Car class, we treat front-view and side-view objects as two different classes. Together with the \textit{Background} class and the \textit{DoNotCare} class, 6 classes are predicted.  We build the graph using $r=1.6m$, and downsample the point cloud by a voxel size of 0.4 meters in training and 0.2 meters in inference.  $MLP_f$ and $MLP_g$ are both of sizes $(256, 256)$. For the vertex state initialization, we set $r_0=0.4m$. We use a an $MLP$ of $(32, 64, 128, 256, 512)$ for embedding and an $MLP$ of $(256, 256)$ to process the aggregated feature.
We set $T_h = 0.2$ in NMS. 

We train the proposed GNN end-to-end with a batch size of 4. The loss weights are $\alpha=0.1$, $\beta=10$, $\gamma=5e-7$. We use stochastic gradient descent (SGD) with a stair-case learning-rate decay. For Car, we use an initial learning rate of $0.125$ and a decay rate of $0.1$ every $400K$ steps. We trained the network for $1400K$ steps. For Pedestrian and Cyclist, we use an initial learning rate of $0.32$ and a decay rate of $0.25$ every $400K$ steps. We trained it for $1000K$ steps. 

\subsection{Data Augmentation}
To prevent overfitting, we perform data augmentation on the training data. Unlike many approaches \cite{SECOND}\cite{PointPillars}\cite{PointRCNN}\cite{STD} that use sophisticated techniques to create new ground truth boxes, we choose a simple scheme of global rotation, global flipping, box translation and vertex jitter. During training, we randomly rotate the point cloud by yaw $\Delta\theta \sim
 \mathcal{N}(0, \pi/8)$ and then flip the $x$-axis by a probability of $0.5$.  After that, each box and points within $110\%$ size of the box randomly shift by $(\Delta x \sim
 \mathcal{N}(0, 3), \Delta y = 0, \Delta z \sim
 \mathcal{N}(0, 3))$. We use a $10\%$ larger box to select the points to prevent cutting the object. During the translation, we check and avoid collisions among boxes, or between background points and boxes.  During graph construction, we use a random voxel downsample to induce vertex jitter.

\subsubsection{Results}\label{results}
We have submitted our results to the KITTI 3D object detection benchmark and the Bird's Eye View (BEV) object detection benchmark. In Table \ref{3D_table} and Table \ref{BEV_table}, we compare our results with the existing literature. The KITTI dataset evaluates the Average Precision (AP) on three difficulty levels: Easy, Moderate, and Hard. Our approach achieves the leading results on the Car detection of Easy and Moderate level and also the Cyclist detection of Moderate and Hard level. Remarkably, on the Easy level BEV Car detection, we surpass the previous state-of-the-art approach by 3.45. Also, we outperform fusion-based algorithms in all categories except for Pedestrian detection.  In Figure \ref{fig:quality}, we provide qualitative detection results on all categories. The results on both the camera image and the point cloud can be visualized. It must be noted that our approach uses \textit{only} the point cloud data. The camera images are purely used for visual inspection since the \textit{test} dataset does not provide ground truth labels. As shown in Figure \ref{fig:quality}, our approach still detects Pedestrian reasonably well despite not achieving the top score. One likely reason why Pedestrian detection is not as good as that for Car and Cyclist is that the vertices are not dense enough to achieve more accurate bounding boxes. 

\subsection{Ablation Study}
For the ablation study, we follow the standard practice \cite{PointPillars}\cite{STD}\cite{Point_RCNN} and split the training samples into a training split of 3712 samples and a validation split of 3769 samples. We use the training split to train the network and evaluate its accuracy on the validation split. We follow the same protocol and assess the accuracy by AP. Unless explicitly modified for a controlled experiment, the network configuration and training parameters are the same as those in the previous section.  We focus on the detection of Car because of its dominant presence in the dataset.

\begin{table}
\centering
\resizebox{\linewidth}{!}{
\begin{tabular}{c|ccc|ccc|ccc}
\hline
 & Box   & Box   & Auto                     & \multicolumn{3}{c|}{BEV AP (Car)} & \multicolumn{3}{c}{3D AP (Car)} \\
 & Merge & Score & Reg.                     & Easy     & Moderate    & Hard    & Easy     & Moderate    & Hard    \\ \hline\hline
1 & -    &    -   &     -                     & 89.11    & 87.14       & 86.18   & 85.46    & 76.80       & 74.89   \\
2 & -    &   -    & $\checkmark$                        & 89.03    & 87.43       & 86.39   & 85.58    & 76.98       & 75.69   \\  
3 & $\checkmark$     &   -    & {$\checkmark$} & 89.33    & 87.83       & 86.63   & 86.59    & 77.49       & 76.35   \\
4 & -    & $\checkmark$     & {$\checkmark$} & 89.60    & 88.02       & 86.97   & 87.40    & 77.90       & 76.75   \\
5 & $\checkmark$     & $\checkmark$     &     -                     & 90.03    & 88.27       & 87.12   & 88.16    & 78.40       & 77.49   \\
6 & $\checkmark$     & $\checkmark$     & $\checkmark$                        & 89.82    & 88.31       & 87.16   & 87.89    & 78.34       & 77.38   \\ \hline
\end{tabular}
}
\vspace*{0mm}
 \caption{Ablation study on the \textit{val.} split of KITTI data. }
\label{design} 
\vspace{-2.5mm}
\end{table}

\noindent \textbf{Box merging and scoring.} In Table \ref{design}, we compare the object detection accuracy with and without box merging and scoring. For the test without box merging, we modify line 11 in Algorithm \ref{nms}. Instead of taking the $median$ bounding box, we directly take the bounding box with the highest classification score as in standard NMS.  For the test without box scoring, we modify lines 12 and 13 in Algorithm \ref{nms} to set the highest classification score as the box score. For the test without box merging or scoring, we modify lines 11, 12, and 13, which essentially leads to standard NMS. Row 2 of Table \ref{design} shows a baseline implementation that uses standard NMS with the auto-registration mechanism. As shown in Row 3 and Row 4 of Table \ref{design}, both box merging and box scoring outperform the baseline. When combined, as shown in Row 6 of the table,  they further outperform the individual accuracy in every category. Similarly, when not using auto-registration, box merging and box scoring (Row 5) also achieve higher accuracy than standard NMS (Row 1). These results demonstrate the effectiveness of the box scoring and merging. 

\noindent \textbf{Auto-registration mechanism.} Table \ref{design} also shows the accuracy improvement from the auto-registration mechanism. As shown in Row 2, by using auto-registration alone, we also surpass the baseline without auto-registration (Row 1) on every category of 3D detection and the moderate and hard categories of BEV detection. The performance on the easy category of BEV detection decreases slightly but remains close. Combining the auto-registration mechanism with box merging and scoring (Row 6), we achieve higher accuracy than using the auto-registration alone (Row 2). However, the combination of all three modules (Row 6) does not outperform box merging and score (Row 5). We hypothesize that the regularization likely needs to be tuned after adding the auto-registration branch. 

\begin{figure}
\begin{center}
\includegraphics[width=\linewidth]{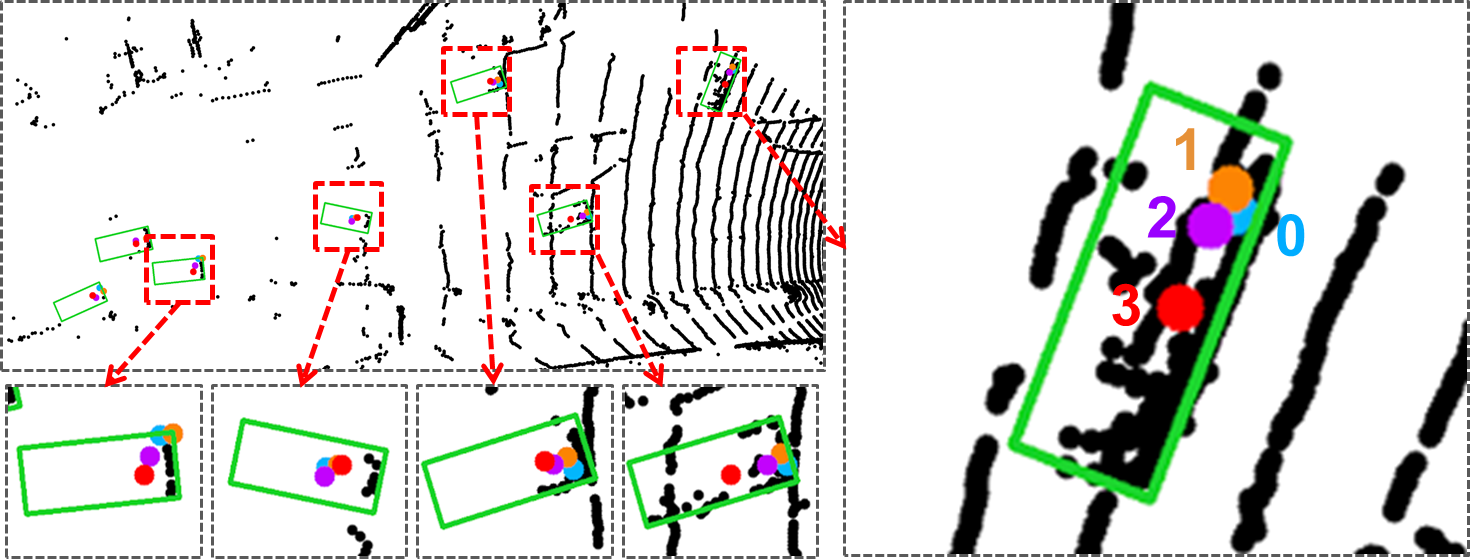}
\end{center}
  \caption{An example from the \textit{val.} split showing the vertex locations with added offsets.  The blue dot indicates the original position of the vertices. The orange, purple, and red dots indicate the original position with added offsets from the first, the second, and the third graph neural network iterations. Best viewed in color. }
\label{fig:auto}
\end{figure}

We further investigate the auto-registration mechanism by visualizing the offset $\Delta x$ in Equation \ref{eq:gnn_invariance}. We extract $\Delta x$ from different GNN iterations and add them to the vertex position. Figure \ref{fig:auto} shows the vertices that output detection results and their positions with added offsets. We observe that the vertex positions with added offsets move towards the center of vehicles. We see such behaviors regardless of the original vertex position. In other words, when the GNN gets deeper, the relative coordinates of the neighbor vertices depend less on the center vertex position but more on the property of the point cloud. The offset $\Delta x$ cancels the translation of the center vertex, and thus reduces the sensitivity to the vertex translation. These qualitative results demonstrate that Equation \ref{eq:gnn_invariance} helps to reduce the translation variance of vertex positions 

\begin{table}[]
\centering
\resizebox{\linewidth}{!}{
\begin{tabular}{c|ccc|ccc}
\hline
Number of                  & \multicolumn{3}{c|}{BEV AP (Car)} & \multicolumn{3}{c}{3D AP (Car)} \\
iterations                     & Easy     & Moderate    & Hard    & Easy     & Moderate    & Hard    \\ \hline\hline
T = 0                        & 87.24    & 77.39       & 75.84   & 73.90    & 64.42       & 59.91   \\
T = 1 & 89.83    & 87.67       & 86.30   & 88.00    & 77.89       & 76.14   \\
T = 2 & 90.00    & 88.37       & 87.22   & 88.34    & 78.51       & 77.67   \\ \hline
T = 3 & 89.82    & 88.31       & 87.16   & 87.89    & 78.34       & 77.38   \\ \hline
\end{tabular}
}
\vspace*{0mm}
\caption{Average precision on the KITTI \textit{val.} split using different number of Point-GNN iterations.}
\label{layers}
\vspace*{-2mm}
\end{table}

\noindent \textbf{Point-GNN iterations.}
Our Point-GNN refine the vertex states iteratively. In Table \ref{layers}, we study the impact of the number of iterations on the detection accuracy. We train Point-GNNs with $T = 1$, $T=2$, and compare them with $T=3$, which is the configuration in Section \ref{results}.  Additionally, we train a detector using the initial vertex state directly without any Point-GNN iteration.  As shown in Table \ref{layers}, the initial vertex state alone achieves the lowest accuracy since it only has a small receptive field around the vertex. Without Point-GNN iterations, the local information cannot flow along the graph edges, and therefore its receptive field cannot expand. Even with a single Point-GNN iteration $T=1$, the accuracy improves significantly. $T=2$ has higher accuracy than $T=3$, which is likely due to the training difficulty when the neural network goes deeper. 

\noindent \textbf{Running-time analysis.} 
The speed of the detection algorithm is important for real-time applications such as autonomous driving. However, multiple factors affect the running time, including algorithm architecture, code optimization and hardware resource. Furthermore, optimizing the implementation is not the focus of this work. However, a breakdown of the current inference time helps with future optimization.  Our implementation is written in Python and uses Tensorflow for GPU computation. We measure the inference time on a desktop with Xeon E5-1630 CPU and GTX 1070 GPU.  The average processing time for one sample in the validation split is 643ms. Reading the dataset and running the calibration takes 11.0\% time (70ms). Creating the graph representation consumes 18.9\% time (121ms). The inference of the GNN takes 56.4\% time (363ms). Box merging and scoring take 13.1\% time (84ms).

\begin{table}[]
\centering
\resizebox{\linewidth}{!}{
\begin{tabular}{c|ccc|ccc}
\hline
Number of                 & \multicolumn{3}{c|}{BEV AP (Car)} & \multicolumn{3}{c}{3D AP (Car)} \\
scanning line             & Easy     & Moderate    & Hard    & Easy     & Moderate    & Hard    \\ \hline\hline
64                        & 89.82    & 88.31       & 87.16   & 87.89    & 78.34       & 77.38   \\
32 & 89.62    & 79.84       & 78.77   & 85.31    & 69.02       & 67.68   \\
16 & 86.56    & 61.69       & 60.57   & 66.67    & 50.23       & 48.29   \\
8  & 49.72    & 34.05       & 32.88   & 26.88    & 21.00       & 19.53   \\ \hline
\end{tabular}
}
\vspace*{0mm}
\caption{Average precision on downsampled KITTI \textit{val.} split.}
\label{scan_lines}
\vspace*{-2mm}
\end{table}

\noindent \textbf{Robustness on LiDAR sparsity.}
The KITTI dataset collects point cloud data using a 64-scanning-line LiDAR. Such a high-density LiDAR usually leads to a high cost. Therefore, it is of interest to investigate the object detection performance in a less dense point cloud. To mimic a LiDAR system with fewer scanning lines, we downsample the scanning lines in the KITTI validation dataset. Because KITTI gives the point cloud without the scanning line information, we use k-means to cluster the elevation angles of points into 64 clusters, where each cluster represents a LiDAR scanning line. We then downsample the point cloud to 32, 16, 8 scanning lines by skipping scanning lines in between. Our test results on the downsampled KITTI validation split are shown in Table \ref{scan_lines}. The accuracy for the moderate and hard levels drops fast with downsampled data, while the detection for the easy level data maintains a reasonable accuracy until it is downsampled to 8 scanning lines. This is because that the easy level objects are mostly close to the LiDAR, and thus have a dense point cloud even if the number of scanning lines drops. 

\section{Conclusion}
We have presented a graph neural network, named Point-GNN, to detect 3D objects from a graph representation of the point cloud. By using a graph representation, we encode the point cloud compactly without mapping to a grid or sampling and grouping repeatedly. Our Point-GNN achieves the leading accuracy in both the 3D and Bird's Eye View object detection of the KITTI benchmark. Our experiments show the proposed auto-registration mechanism reduces transition variance, and the box merging and scoring operation improves the detection accuracy. In the future, we plan to optimize the inference speed and also fuse the inputs from other sensors.

{\small
\bibliographystyle{ieee_fullname}
\bibliography{egbib}
}

\end{document}